\documentclass[times,twocolumn,final]{elsarticle}

\usepackage[colorlinks,linkcolor=red,anchorcolor=blue, citecolor=blue]{hyperref}
\usepackage{multirow}
\usepackage[T1]{fontenc}
\usepackage{graphicx}
\usepackage{cite}
\usepackage{subfigure}
\usepackage{tabularx}
\usepackage{float}
\usepackage{amsmath} 
\usepackage{amssymb}
\usepackage{booktabs}
\usepackage{colortbl}
\usepackage{todonotes}
\usepackage{color}
\usepackage{overpic}
\usepackage{amsmath}
\usepackage{subfigure}
\usepackage{bm}
\usepackage{bbding}
\usepackage{makecell}
\usepackage{array}
\usepackage{ulem}

\usepackage{medima}
\usepackage{framed,multirow}

\usepackage{amssymb}
\usepackage{latexsym}

\usepackage{url}
\usepackage{xcolor}

\usepackage{hyperref}

\definecolor{newcolor}{rgb}{.8,.349,.1}

\journal{Medical Image Analysis}

\begin{document}

\verso{Jiang \textit{et~al.}}

\begin{frontmatter}

\title{A 3D SAM-Based Progressive Prompting Framework for Multi-Task Segmentation of Radiotherapy-induced Normal Tissue Injuries in Limited-Data Settings}%

\author[1]{Caiwen Jiang }
\author[1]{Lei Zeng}
\author[1]{Wei Liu}



\address[1]{Department of Radiation Oncology, Mayo Clinic, Phoenix, Arizona, USA}


\begin{abstract}
Radiotherapy-induced normal tissue injury is a clinically important complication, and accurate segmentation of injury regions from medical images could facilitate disease assessment, treatment planning, and longitudinal monitoring. However, automatic segmentation of these lesions remains largely unexplored because of multiple challenges, including scarce high-quality voxel-level annotations and their extremely small, sparse presentation in 3D images. To address this gap, we curate a dedicated head-and-neck radiotherapy–induced normal tissue injury dataset covering multiple manifestations, including osteoradionecrosis (ORN), cerebral edema (CE), cerebral radiation necrosis (CRN), and develop a SAM-based promptable segmentation framework for multi-task segmentation of small injury regions in a limited-data setting. Specifically, the framework adopts a progressive prompting strategy built upon three complementary types of prompts, moving from task conditioning to coarse localization and iterative refinement: 1) text prompts, which encode task-specific and clinically relevant information to guide adaptation of the pretrained 3D SAM backbone, thereby enabling task-conditioned segmentation of multiple injury manifestations while leveraging large-scale pretraining priors; 2) dose-guided box prompts, which are derived by identifying high-dose regions from radiotherapy dose distributions to constrain the search space and facilitate coarse localization of tiny and sparse lesions; and 3) click prompts, which are automatically generated via a click simulation mechanism and incorporated together with a small-target focus loss to iteratively refine local predictions and improve lesion boundary delineation. Extensive experiments across osteoradionecrosis (ORN), cerebral edema (CE), cerebral radiation necrosis (CRN) demonstrate that our method enables reliable segmentation of diverse head-and-neck radiotherapy–induced normal tissue injuries and outperforms state-of-the-art methods, highlighting its potential for automated injury assessment and longitudinal monitoring.
\end{abstract}

\end{frontmatter}

\section{Introduction}
\label{sec:introduction}

Radiotherapy is a cornerstone of modern cancer treatment that eradicates tumour cells through ionizing radiation, but the unavoidable exposure of surrounding normal tissues can also lead to adverse effects \citep{delaney2005role,bentzen2010quantec}. Among these, radiotherapy-induced normal tissue injuries are clinically important complications that may substantially impair patient quality of life and complicate subsequent clinical management \citep{bentzen2010quantec,marks2010use}. Accurate delineation of injury regions from medical images is therefore essential for disease assessment, treatment planning, treatment response evaluation, and longitudinal monitoring. However, manual delineation of injury regions in 3D medical images is labor-intensive, time-consuming, highly dependent on clinical expertise, and prone to inter-observer variability \citep{vinod2016interobserver}. These limitations underscore the need for automatic segmentation algorithms that enable more efficient, objective, and scalable image-based evaluation \citep{ronneberger2015u}.

In recent years, deep learning-based medical image segmentation methods have achieved remarkable success across a broad range of applications \citep{isensee2021nnunet,kirillov2023segmentanything,ma2024medsam}. Nevertheless, segmentation of radiotherapy-induced normal tissue injuries remains largely unexplored, and general-purpose segmentation algorithms often perform suboptimally in this setting. This can be attributed to three major task-specific challenges. First, radiotherapy-induced normal tissue injuries are relatively uncommon in routine clinical imaging, while voxel-level annotation of injury regions is labor-intensive and highly dependent on clinical expertise, resulting in pronounced data scarcity. Second, injury regions are often extremely small and sparsely distributed within 3D images, leading to severe target-background imbalance and making reliable localization and precise delineation inherently difficult. Third, radiotherapy may induce multiple injury manifestations, either concurrently or sequentially, rather than a single isolated outcome. Consequently, training separate models for different manifestations is inflexible and fails to exploit their intrinsic relationships, whereas joint segmentation within a unified framework remains highly challenging. These difficulties are further exacerbated under limited-data conditions.

In this context, recent advances in promptable segmentation have opened new opportunities for medical image analysis under limited-data conditions. In particular, the segment anything model (SAM) and its 3D extensions \citep{kirillov2023segmentanything,ma2024medsam,zhang2024sammed3d} are especially attractive for limited-data segmentation because they combine transferable priors from large-scale pretraining with a promptable segmentation paradigm that allows task-relevant guidance to be injected explicitly. This provides a flexible foundation for tasks in which task specification, spatial prior information, and local refinement all play important roles. However, directly fine-tuning SAM with a generic prompting scheme is still unlikely to fully address the challenges of radiotherapy-induced normal tissue injury segmentation. Beyond limited data, this task also requires effective modeling of multiple injury manifestations, reliable localization of tiny and sparse lesions, and fine-grained delineation of ambiguous boundaries. These considerations motivate the development of a more task-tailored prompting strategy that can introduce complementary guidance throughout the segmentation process.

\begin{figure}[!t]
\setlength{\belowcaptionskip}{-0.4cm}
\centering
\begin{overpic}[width=1\linewidth]{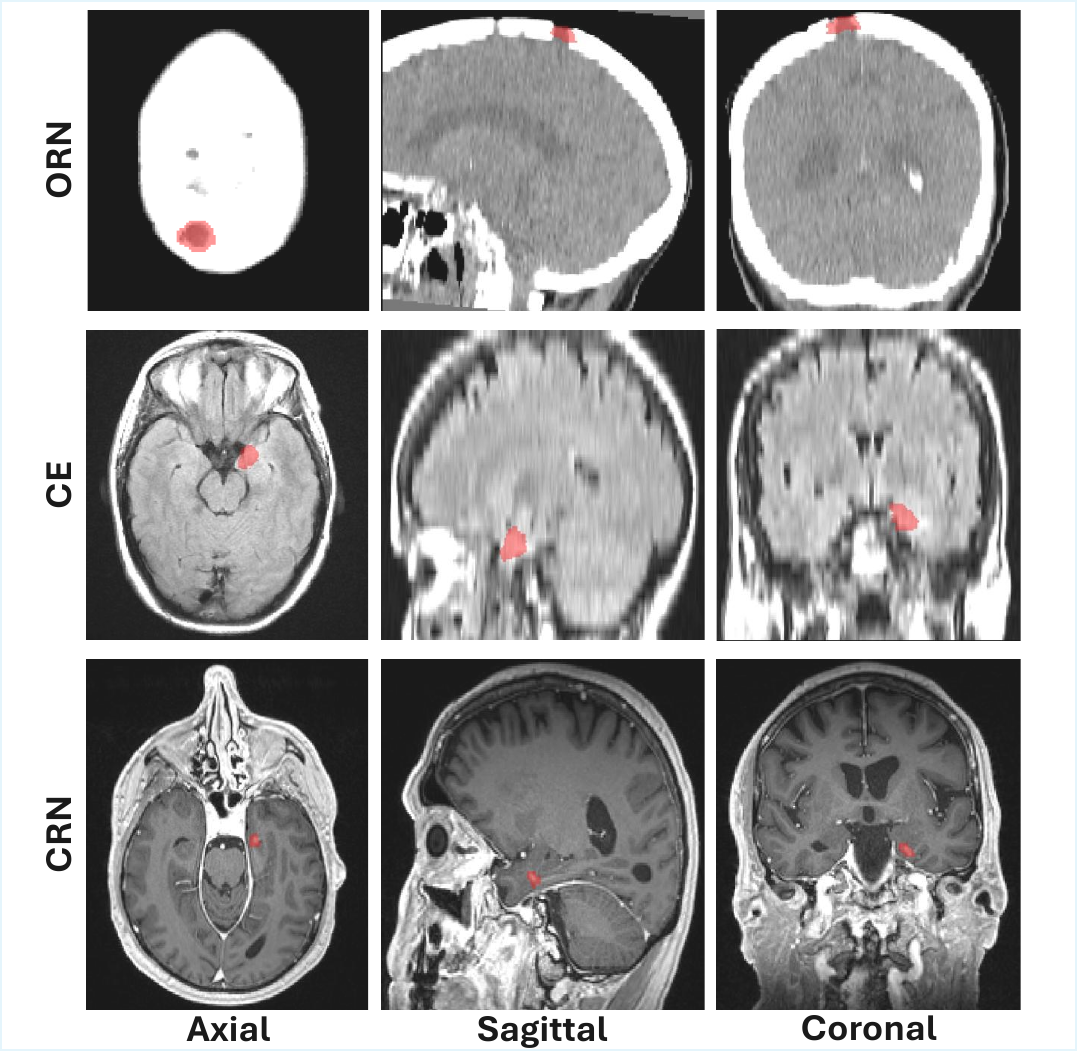}

    \end{overpic}
    \vspace{-1mm}
\centering
 \setlength{\abovecaptionskip}{0.1cm}
 
\caption{Representative examples of radiotherapy-induced normal tissue injuries in the curated dataset.
Three injury types, osteoradionecrosis (ORN), cerebral edema (CE), and cerebral radiation necrosis (CRN), are displayed in axial, sagittal, and coronal views. ORN is shown on CT, CE on T2-FLAIR MR images, and CRN on contrast-enhanced T1-weighted MR images. Lesion regions are highlighted in red.}
\label{example}
\end{figure}

To this end, we propose a 3D SAM-based progressive prompting framework for multi-task segmentation of radiotherapy-induced normal tissue injuries, built upon a complementary multi-prompt design that explicitly incorporates radiotherapy-specific priors and injury-aware guidance. Concretely, the framework first uses text prompts to encode task-specific and clinically relevant information, guiding fine-tuning of the pretrained 3D SAM backbone for multi-task radiotherapy injury segmentation. This allows multiple injury manifestations to be modeled within a unified framework while retaining the benefits of large-scale pretraining priors. It then introduces dose-guided box prompts derived from high-dose regions in radiotherapy dose distributions, exploiting the clinical prior that injury generally occurs within highly irradiated areas to constrain the search space and facilitate target localization. Finally, click prompts are automatically generated through a click simulation mechanism based on discrepancies between predictions and ground truth, yielding both foreground-aware and background-aware guidance, and are coupled with a small-target focus loss to iteratively refine local predictions and improve lesion boundary delineation. Together, these three prompts provide complementary guidance in a progressive manner, enabling a unified segmentation pipeline that moves from task conditioning to lesion localization and fine-grained refinement.

The main contributions of this work can be summarized as follows:
\begin{enumerate}
    \item
We investigate automatic segmentation of radiotherapy-induced normal tissue injuries, a clinically important yet underexplored task, and curate a dedicated dataset covering multiple injury manifestations, illustrated in Fig.~\ref{example}, to support this study.

     \item
    To address this task, we propose a 3D SAM-based progressive prompting framework for multi-task segmentation of radiotherapy-induced normal tissue injuries in limited-data settings, enabling multiple injury manifestations to be handled within a unified model.

    \item
    Within this framework, we design a complementary multi-prompt strategy that progressively integrates text prompts for task conditioning, dose-guided box prompts for coarse localization, and automatically simulated click prompts together with a small-target focus loss for iterative refinement of small and sparse injury regions.

    \item
Extensive experiments demonstrate the feasibility and effectiveness of SAM-based automatic segmentation for radiotherapy-induced normal tissue injuries. The proposed method achieves reliable performance across ORN, CE, and CRN, consistently outperforming state-of-the-art segmentation methods and highlighting its value as a powerful tool for future injury analysis.
\end{enumerate}

The rest of this paper is organized as follows. We will first introduce related works in Section 2, and then describe the details of our proposed approach in Section 3. Next, we will give implementation details and experimental results in Section 4, and provide some extended discussion in Section 5. Finally, we will summarize our work in Section 6.

\section{Related Work}
\subsection{Deep Learning for Radiotherapy Image Analysis}
Deep learning has become a major driving force in radiotherapy image analysis, with applications spanning target and organ-at-risk delineation, treatment planning, dose prediction, adaptive radiotherapy, and outcome modeling \citep{lagedamon2024dose,lemus2024adaptive}. Among these areas, automatic image segmentation has emerged as one of the most clinically important and extensively studied topics. For example, Gibbons \textit{et al.} evaluate deep learning-based auto-segmentation of critical organs-at-risk (OARs) for routine radiation therapy contouring in comparison with atlas-based methods \citep{gibbons2023clinical}. In a different radiotherapy setting, Cubero \textit{et al.} develop a deep learning approach for head-and-neck OAR segmentation on cone-beam CT, targeting adaptive radiotherapy workflows rather than conventional planning CT segmentation \citep{cubero2025cbct}. In addition to normal-structure delineation, deep learning has also been applied to the definition of treatment targets. Wu \textit{et al.} propose a deep learning method for automatic clinical target volume (CTV) delineation in cervical cancer radiotherapy, demonstrating that target auto-contouring can improve efficiency without compromising contour quality \citep{wu2025ctv}.

However, despite these advances, radiotherapy segmentation research has remained focused largely on organs-at-risk and treatment targets, whereas treatment-induced tissue injuries have received much less attention from a lesion segmentation perspective. Existing imaging studies of radiotherapy-induced normal tissue injuries have predominantly concentrated on risk prediction and outcome modeling. For example, Reber \textit{et al.} investigate mandibular ORN prediction by comparing dose-based conventional machine-learning and deep-learning models \citep{reber2023orn}, while Barua \textit{et al.} explore early ORN prediction using longitudinal CT radiomics kinetics derived from sequential mandibular imaging \citep{barua2021ct}. Although such prediction-oriented studies are valuable for identifying patients or subregions at elevated risk, they do not directly characterize the spatial extent, morphology, or burden of established lesions, thereby limiting their utility for detailed toxicity assessment, longitudinal monitoring, and downstream image-based analysis. To our knowledge, published studies on automatic segmentation of radiotherapy-related post-treatment changes remain extremely limited, and the few available examples appear to come largely from a single research group. In particular, this group combines Mask R-CNN and DeepMedic to segment post-radiosurgery brain edema in meningioma patients and radiation-induced changes following stereotactic radiosurgery for arteriovenous malformations \citep{yang2025edema,ho2025ric}. While these studies provide useful proof of feasibility, they are better viewed as early application-oriented attempts based on existing general-purpose segmentation models. More importantly, each of them is tailored to a single injury manifestation in a specific clinical scenario. In clinical practice, however, radiotherapy-induced normal tissue injuries may arise in multiple forms across different settings. Consequently, existing studies remain insufficient to address the broader challenge of unified segmentation across diverse injury manifestations, especially when lesions are small, sparse, and spatially scattered.

\subsection{Medical Image Segmentation with Limited Annotated Data}

Although deep learning has achieved strong performance in medical image segmentation, it often relies on sufficiently large datasets with dense expert annotations. In practice, however, voxel-level labeling is costly and time-consuming, particularly for lesion segmentation tasks that require careful slice-by-slice delineation by experienced clinicians. As a result, limited annotated data remains a persistent challenge, and existing studies mainly address this problem through semi-supervised learning, data generation-based augmentation, and task-specific adaptation of pre-trained general models. Among these, semi-supervised learning reduces reliance on dense annotations by combining a small labeled set with a larger pool of unlabeled images. For example, Liu \textit{et al.} develop a semi-supervised left atrium segmentation framework that combines a segmentation network with a classification network and leverages unlabeled images through a contrastive consistency loss on class vectors \citep{liu2022la}. Vorontsov \textit{et al.} propose a tumor segmentation framework that combines a few pixel-level annotations with weak image-level tumor labels, using the former for direct segmentation supervision and the latter to support image-to-image translation between healthy and diseased domains as an auxiliary objective \citep{vorontsov2022annotation}. Overall, these methods reduce dependence on dense annotations, but their effectiveness still depends strongly on the quality of unlabeled or weakly labeled data, and even small errors in pseudo-labels, coarse supervision, or mismatches between labeled and unlabeled data distributions can substantially affect representation learning.

Beyond leveraging unlabeled data, data generation-based augmentation alleviates annotation scarcity by using generative models to synthesize medical images together with their corresponding masks. For example, Zhang \textit{et al.} introduce a generative framework for medical image segmentation that synthesizes paired masks and medical images as auxiliary training data, with segmentation and generation jointly optimized to improve learning under limited annotations \citep{zhang2025genseg}. A related example is provided by Pandey \textit{et al.}, who propose a two-stage generative adversarial framework for segmentation augmentation, in which masks are generated first and corresponding images are then synthesized to expand the training set \citep{pandey2020twostage}. These methods provide a complementary way to mitigate annotation scarcity by enriching the available training distribution. However, their effectiveness still depends heavily on the realism, diversity, and task relevance of the generated samples. This limitation is particularly pronounced in relatively rare and heterogeneous scenarios, where scarce underlying data make it difficult to generate sufficiently realistic and representative training samples.

More recent studies increasingly address limited-data segmentation through task-specific adaptation of pre-trained general models, particularly SAM-based models such as SAM, MedSAM, and SAM-Med3D \citep{kirillov2023segmentanything,ma2024medsam,zhang2024sammed3d}. The central idea is to first learn transferable visual or segmentation priors from large-scale datasets and then adapt the pre-trained model to a downstream task with limited annotations. For example, Chen \textit{et al.} adapt SAM to volumetric medical segmentation in MA-SAM through parameter-efficient fine-tuning with 3D adapters \citep{chen2024masam}. Guo \textit{et al.} fine-tune SAM in ClickSAM for ultrasound image segmentation using click prompts \citep{guo2024clicksam}. Compared with semi-supervised learning and data generation, this strategy more directly reduces dependence on task-specific labeled data by leveraging priors acquired during pretraining. However, its success hinges on effective task-specific adaptation strategies, as directly applying a general pre-trained model is often insufficient for challenging lesion segmentation tasks, especially when the targets are small, sparse, and heterogeneous, such as radiotherapy-induced injuries.

\subsection{Promptable Learning in Medical Image Analysis}
Promptable learning uses external prompts, including spatial cues such as points and boxes, as well as semantic cues such as text descriptions or task identifiers, to steer model behavior, thereby enabling more controllable and task-aware inference than conventional task-specific models with fixed input-output mappings. Owing to this flexibility, it has emerged as a promising paradigm for introducing task-specific guidance into model inference, with growing applications in medical image analysis tasks such as localization, detection, and multimodal understanding. 

Recent studies have explored this paradigm in various medical image analysis settings. Denner \textit{et al.} explore visual prompt engineering for radiology by embedding explicit visual markers, such as arrows, circles, and contours, into images to guide model attention in lung nodule malignancy classification \citep{denner2024visual}. Deng \textit{et al.} develop GK-MVLP, which constructs grounded medical knowledge prompts and aligns them with anatomical regions to improve chest X-ray disease localization, report generation, and medical visual question answering \citep{deng2024gkmvlp}.  Lai \textit{et al.} propose CARZero, which employs large language models to generate prompt templates from radiology reports and couples them with cross-attention alignment for radiology zero-shot classification \citep{lai2024carzero}. Bie \textit{et al.} propose XCoOp, an explainable prompt learning framework for computer-aided diagnosis that aligns images, learnable prompts, and clinical concept-driven prompts at multiple granularities to incorporate medical knowledge into diagnosis \citep{bie2024xcoop}. Collectively, these studies demonstrate the ability of promptable learning to introduce explicit task-specific guidance into model inference, enabling more flexible and controllable adaptation across diverse medical image analysis scenarios. Inspired by these advances, we explore promptable strategies as a task-specific adaptation mechanism to better tailor the general segmentation model to the challenging task of radiotherapy-induced injury segmentation.

\section{Method}
\begin{figure*}[!t]
\setlength{\belowcaptionskip}{-0.4cm}
\centering
\begin{overpic}[width=1\linewidth]{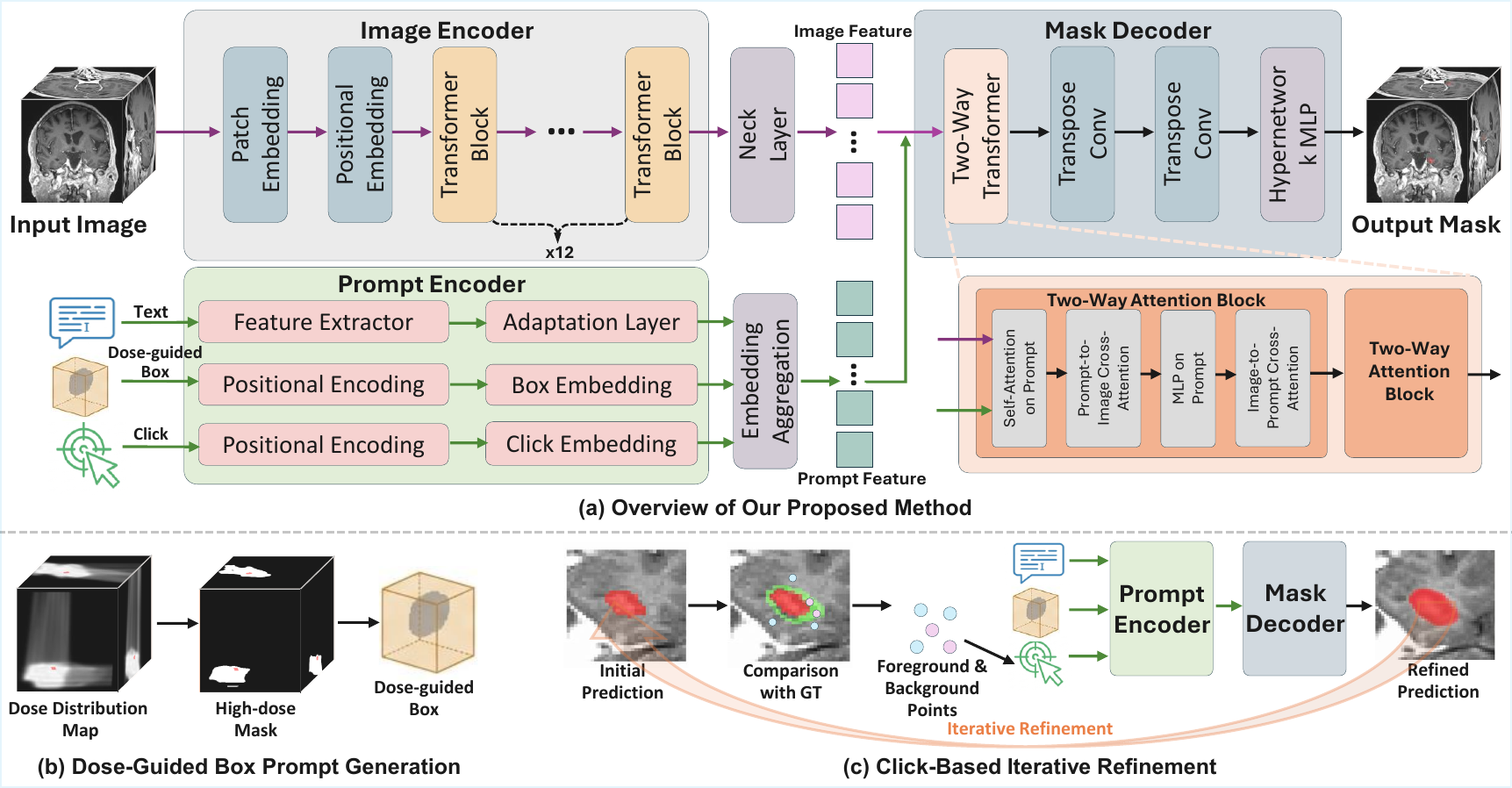}

    \end{overpic}
    \vspace{-1mm}
\centering
 \setlength{\abovecaptionskip}{0.1cm}
 
\caption{Schematic illustration of the proposed progressive prompting framework. (a) Overall architecture of the proposed method, including the image encoder, prompt encoder, and mask decoder. (b) Dose-guided box prompt generation from the high-dose region, with the target region highlighted in red, for coarse lesion localization. (c) Click-based iterative boundary refinement, where click simulation automatically generates foreground and background points to progressively refine the segmentation boundaries.}
\label{framework}
\end{figure*}

As illustrated in Fig.~\ref{framework}(a), the proposed progressive prompting framework consists of three main components, namely the image encoder, prompt encoder, and mask decoder. Given an input 3D medical image, whose modality may vary across different segmentation tasks, the image encoder first extracts image features. Meanwhile, the prompt encoder transforms three types of task-relevant prompts, including text prompts, dose-guided box prompts, and click prompts, into corresponding prompt features that provide progressive guidance for task conditioning, lesion localization, and iterative refinement. The resulting image and prompt features are then jointly fed into the mask decoder to generate the target lesion mask. The details of the proposed method are described in the following subsections.

\subsection{SAM-based Architecture for Leveraging Large-scale Pretrained Priors}
Radiotherapy-induced normal tissue injury segmentation spans different injury types and imaging modalities. This variability makes conventional task-specific segmentation networks less suitable for achieving both robust generalization and flexible adaptation across different tasks, especially under limited-data conditions. To address this challenge, we build our method upon SAM, leveraging its transferable priors learned through large-scale pretraining to enable more effective adaptation under limited task-specific annotations. On this basis, we further develop a progressive promptable segmentation framework with task-tailored prompts for radiotherapy-induced normal tissue injury segmentation.

As shown in Fig.~\ref{framework}(a), the proposed progressive promptable segmentation framework starts with the image encoder to extract image features from the input volume. Specifically, the input 3D image is first partitioned into non-overlapping 3D patches and projected by the \textit{Patch Embedding} layer into latent patch tokens through a 3D convolution with kernel size and stride both set to the patch size. A learnable \textit{Positional Embedding} is then added to these tokens to encode their spatial locations in the volume before feature modeling. The resulting token sequence is subsequently processed by a stack of 12 \textit{Transformer Blocks}. Each block consists of a normalization layer, a multi-head self-attention module, and an MLP block, with residual connections applied after both the attention and MLP operations. In addition, the 12 Transformer Blocks combine window-based \citep{liu2021swin} and global attention \citep{dosovitskiy2021vit} to balance local feature aggregation and long-range dependency modeling, where the 3rd, 6th, 9th, and 12th blocks use global attention, while the remaining blocks use window-based attention. Finally, a \textit{Neck Layer}, composed of consecutive convolutional and normalization operations, transforms the encoder output into compact image features. In parallel, three types of task-relevant prompts, including text prompts, dose-guided box prompts, and click prompts, are encoded by the prompt encoder into prompt features. These prompt features are then integrated with the extracted image features and jointly fed into the mask decoder for prompt-guided mask decoding process. The detailed generation of different prompts and their feature encoding process will be introduced in Section 3.2.

Given the encoded image and prompt features, the mask decoder is responsible for generating the final segmentation mask. Specifically, the image and prompt features are first jointly processed by a \textit{Two-Way Transformer}, in which they are treated as two parallel feature streams and iteratively updated through bidirectional interaction. The \textit{Two-Way Transformer} consists of two stacked \textit{Two-Way Attention Blocks}. Within each block, \textit{Self-Attention on Prompt} is first applied only to the prompt stream to model correlations among different prompt tokens and update their contextual representations, while the image stream remains unchanged. Next, \textit{Prompt-to-Image Cross-Attention} uses the updated prompt features as queries, and the image features as keys and values, allowing the prompt stream to attend to the image stream and incorporate image-aware information. The resulting prompt features are then further processed by an \textit{MLP on Prompt}, which performs token-wise nonlinear transformation on the prompt stream while keeping the image stream unchanged. Finally, \textit{Image-to-Prompt Cross-Attention} takes the image features as queries and the refined prompt features as keys and values, enabling the image stream to attend back to the prompt stream and thereby inject prompt-aware guidance into the image representation. Through this bidirectional interaction, the decoder progressively updates both prompt and image representations, enabling prompt information to guide image feature aggregation while image features also provide feedback to refine prompt-aware representations. After the two-way transformer, the updated image features are progressively upsampled by two \textit{Transpose Conv} layers to recover spatial resolution. Based on the decoder outputs, a \textit{Hypernetwork MLP} further generates dynamic mask coefficients, which are then applied to the upsampled feature maps to produce the final segmentation mask.

Built upon the SAM framework, our method benefits from large-scale pretraining for radiotherapy-induced injury segmentation. The image encoder,  prompt encoder (i.e., the box and click branches), and mask decoder are initialized with weights pretrained on SA-Med3D-140K \citep{zhang2024sammed3d}, a large-scale dataset assembled from 70 public and 24 private datasets and comprising approximately 22,000 3D medical images and 143,000 3D segmentation masks. The framework is then adapted to the target task using the limited radiotherapy-induced injury data collected in this study, with the image encoder kept frozen while the modified prompt encoder and mask decoder are updated. This adaptation strategy allows the framework to retain generalizable knowledge from large-scale pretraining while acquiring task-specific capability for radiotherapy-induced injury segmentation.

\subsection{Progressive Prompting Strategy}
While the SAM-based architecture provides a strong foundation for leveraging large-scale pretrained priors, effective adaptation to radiotherapy-induced normal tissue injury segmentation still requires task-specific guidance. This is because the task involves substantial heterogeneity across injury types and imaging modalities, as well as extremely small and sparse lesions that are difficult to localize and delineate under limited-data conditions. To address these challenges, we propose a progressive prompting strategy, in which three complementary prompts, i.e., text prompts, dose-guided box prompts, and click prompts, provide hierarchical guidance from task-level conditioning to coarse spatial localization and further to local boundary refinement. These prompts are considered progressive because their guidance becomes increasingly specific in both functional role and granularity, moving from semantic understanding to lesion localization and finally to boundary refinement. The three prompt types are detailed in the following subsections.

\begin{figure}[!t]
\setlength{\belowcaptionskip}{-0.4cm}
\centering
\begin{overpic}[width=1\linewidth]{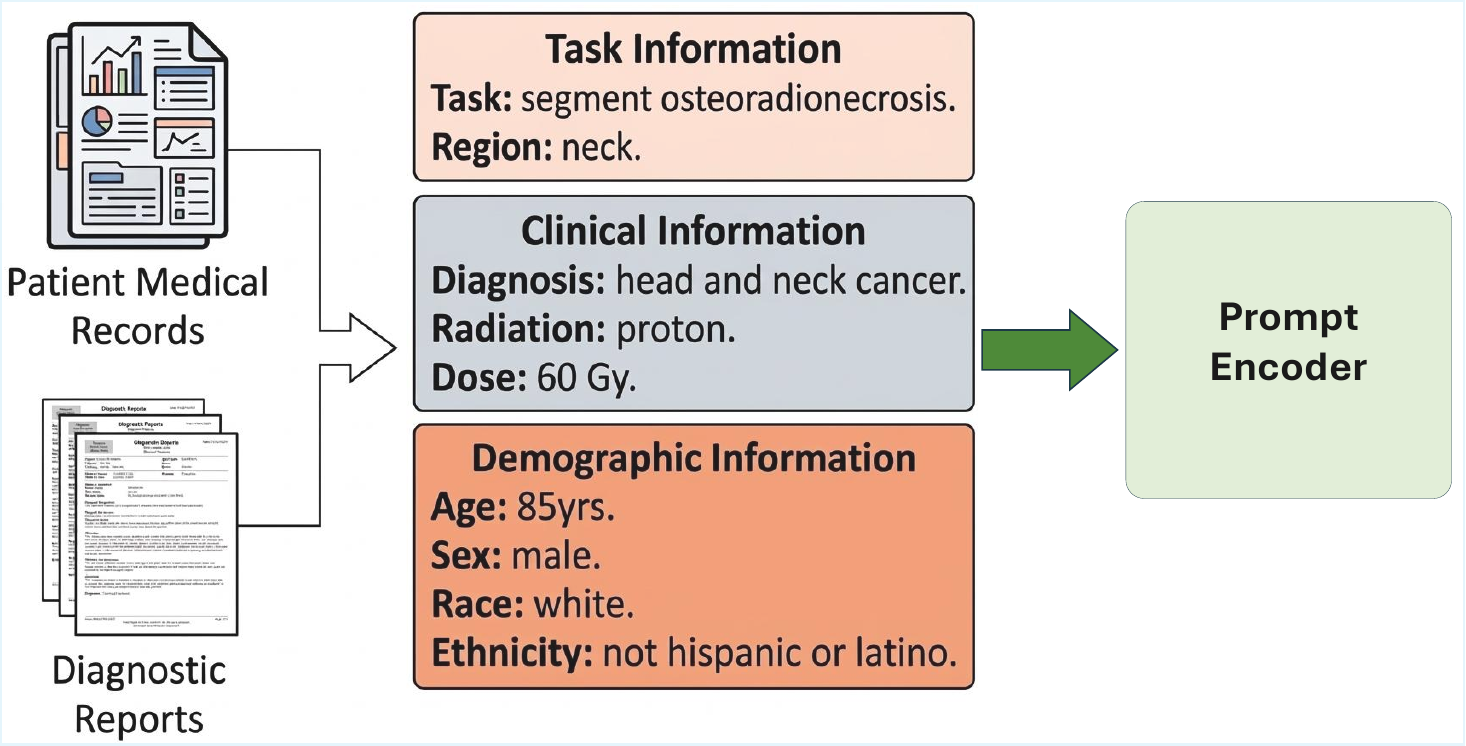}

    \end{overpic}
    \vspace{-1mm}
\centering
 \setlength{\abovecaptionskip}{0.1cm}
 
\caption{Text prompt construction from patient medical records and diagnostic reports, including task, clinical, and demographic information.}
\label{text-prompt}
\end{figure}

\subsubsection{Text Prompt for Task Conditioning}

To provide explicit semantic guidance for task conditioning, we incorporate text prompts into the proposed framework. Specifically, as shown in Fig.~\ref{text-prompt}, three types of textual information are collected from patient medical records and diagnostic reports, including task information, clinical information, and demographic information, which are jointly used as the input text prompt. Here, task information defines the target injury segmentation task, whereas clinical information and demographic information provide complementary patient-specific context. In the text branch of the prompt encoder, the input text prompt is first processed by a pretrained and frozen Bio\_ClinicalBERT \citep{alsentzer2019clinicalbert} model as the \textit{Feature Extractor} to obtain high-level semantic features. Pretrained on biomedical and clinical corpora, this model captures clinically relevant terminology and semantic context from the input text, yielding informative task-related representations. These features are then passed through an \textit{Adaptation Layer} composed of two linear projection layers with a nonlinear activation in between, followed by layer normalization, to project them into the prompt embedding space and generate text prompt tokens for the subsequent radiotherapy-induced injury segmentation task. Through this design, the text prompt provides explicit semantic guidance for task conditioning and equips the model with task-relevant contextual priors for the input case.

\subsubsection{Dose-guided Box Prompts for Coarse Lesion Localization}
Radiotherapy-induced injury regions are often extremely small and sparsely distributed, making direct lesion segmentation highly challenging. To address this issue, we introduce dose-guided box prompts for coarse lesion localization, based on the clinical prior that such injuries usually arise within highly irradiated areas. As shown in Fig.~\ref{framework}(b), we first derive a high-dose mask from the radiotherapy dose map to represent the spatial distribution of highly irradiated regions. Specifically, the maximum dose value $D_{\max}$ is computed from all positive-dose voxels, and a relative threshold of $\tau \times D_{\max}$ is used to extract the high-dose region, where $\tau$ controls the spatial extent of the extracted mask. In our implementation, $\tau$ is set to 0.8. To improve spatial consistency and suppress scattered responses, only the largest connected component is retained. A 3D bounding box is then generated by identifying the minimum and maximum coordinates of this mask along the three spatial dimensions and is used as the dose-guided box prompt. In the box branch of the prompt encoder, the dose-guided bounding box is represented by the coordinates of its two opposite 3D corners. These coordinates are first passed through \textit{Positional Encoding}, which maps the normalized corner coordinates into continuous positional representations to explicitly encode the geometric location of the box. The resulting features are then processed by \textit{Box Embedding}, where two learnable corner-type embeddings are added to distinguish the minimum and maximum corners. In this way, the box branch transforms the coarse spatial extent of the high-dose region into box prompt features that are compatible with the subsequent mask decoding process and provide coarse yet explicit spatial guidance for subsequent lesion segmentation.

\subsubsection{Click Prompts for Iterative Boundary Refinement}
Although dose-guided box prompts provide effective coarse localization, accurate delineation still requires correction of local prediction errors, particularly around ambiguous lesion boundaries. To this end, we introduce click prompts for iterative boundary refinement. As illustrated in Fig.~\ref{framework}(c), click prompts are automatically generated based on the discrepancy between the current prediction and the ground truth. Specifically, false-negative and false-positive regions are first identified by comparing the predicted mask with the ground-truth mask. A click point is then randomly sampled from the union of these error regions. Points sampled from false-negative regions are assigned as positive clicks, while those sampled from false-positive regions are assigned as negative clicks. In this way, each click prompt explicitly indicates where the current prediction should be expanded or suppressed. In the click branch of the prompt encoder, each click prompt is represented by a 3D spatial coordinate and a binary click label. The coordinate is first processed by \textit{Positional Encoding} to transform its spatial location into a continuous positional feature. The resulting feature is then passed to \textit{Click Embedding}, where a learnable type embedding is added according to the click label, enabling positive and negative clicks to be distinguished in the prompt feature space. Through this process, the sampled click prompts are converted into click prompt features that can be jointly used with image and other prompt features in the subsequent mask decoding stage.

The click prompting process is performed iteratively during training. Specifically, the input image, text prompt, and dose-guided box prompt remain fixed, while the click prompt is updated according to the latest prediction error. Starting from the current prediction, a new click prompt is generated from the discrepancy between the prediction and the ground truth, and is then fed back into the prompt encoder and mask decoder to update the segmentation result. This procedure is repeated for multiple rounds, allowing the model to progressively correct local errors and refine lesion boundaries. Notably, click prompts are used only during training, when ground-truth annotations are available for error-based click simulation, and are not used during testing. In this way, click prompting serves as a training-time iterative refinement mechanism that promotes more effective learning of local boundary correction.

\subsection{Small-target Focus Loss}

Radiotherapy-induced injury regions usually occupy only a very small fraction of the whole image volume, making direct whole-volume supervision easily dominated by irrelevant background voxels. To complement the proposed progressive prompting strategy, we further design a small-target focus loss that restricts optimization to the high-dose region of interest, where radiotherapy-induced injuries typically arise, and combines region-overlap supervision with hard-example learning.

Specifically, let $p$ denote the predicted probability map, $y$ the ground-truth (GT) mask, and $r$ the binary high-dose ROI mask derived according to the procedure described in Section 3.2.2. Instead of computing the loss over all voxels, we evaluate the segmentation loss only within voxels satisfying $r(v)=1$, so that optimization is concentrated on the clinically relevant region. Based on this ROI-restricted supervision, the proposed small-target focus loss is defined as
\begin{equation}
\mathcal{L}_{\mathrm{STF}} = \lambda_{\mathrm{1}} \mathcal{L}_{\mathrm{Dice}} + \lambda_{\mathrm{2}} \mathcal{L}_{\mathrm{FT}},
\end{equation}
where $\lambda_{\mathrm{1}}$ and $\lambda_{\mathrm{2}}$ are the balancing weights of the two terms. In our implementation, $\lambda_{\mathrm{Dice}}=0.7$ and $\lambda_{\mathrm{FT}}=0.3$.

The ROI-restricted Dice loss is formulated as
\begin{equation}
\mathcal{L}_{\mathrm{Dice}}
=
1-
\frac{2\sum\limits_{v \in \Omega_r} p(v)y(v)+\epsilon}
{\sum\limits_{v \in \Omega_r} p(v)+\sum\limits_{v \in \Omega_r} y(v)+\epsilon},
\end{equation}
where $\Omega_r=\{v \mid r(v)=1\}$ denotes the set of voxels inside the ROI and $\epsilon$ is a small constant for numerical stability. This term directly optimizes the overlap between the prediction and the GT within the clinically relevant region.

To further emphasize difficult voxels and alleviate the severe foreground-background imbalance within the ROI, we additionally adopt a Focal Tversky loss \citep{abraham2019focaltversky}:
\begin{equation}
\mathcal{L}_{\mathrm{FT}}
=
\left(1-
\frac{TP+\epsilon}{TP+\alpha FP+\beta FN+\epsilon}
\right)^{\gamma},
\end{equation}

\begin{equation}
\begin{aligned}
TP &= \sum\limits_{v \in \Omega_r} p(v)y(v), \\
FP &= \sum\limits_{v \in \Omega_r} p(v)(1-y(v)), \\
FN &= \sum\limits_{v \in \Omega_r} (1-p(v))y(v).
\end{aligned}
\end{equation}
In our implementation, $\alpha=0.5$, $\beta=0.5$, and $\gamma=0.75$. This term increases the contribution of hard-to-segment voxels and thus further improves optimization for small and sparse lesion regions.

Overall, the proposed small-target focus loss facilitates adaptation of the pretrained SAM architecture to radiotherapy-induced injury segmentation by encouraging the network to focus on small injury regions within the high-dose area. It further complements the progressive prompting strategy by strengthening supervision on clinically relevant local regions.

\begin{table*}[!t]
\setlength{\abovecaptionskip}{0.1cm}
\setlength{\belowcaptionskip}{0.1cm}
\centering
\renewcommand\arraystretch{1.25}
\setlength{\tabcolsep}{6pt}
\arrayrulecolor{black}
\caption{Quantitative comparison of the proposed method and five state-of-the-art segmentation methods for radiotherapy-induced injury segmentation, evaluated using Dice, IoU, Precision, Recall, HD95 and ASSD.}
\label{tab:sota_comparison}
\begin{tabular}{l|c|c|c|c|c|c}
\hline
\rowcolor[gray]{0.85}
\textbf{Method} & \textbf{Dice\%$\uparrow$} & \textbf{IoU\%$\uparrow$} & \textbf{Precision\%$\uparrow$} & \textbf{Recall\%$\uparrow$} & \textbf{HD95$\downarrow$} & \textbf{ASSD$\downarrow$} \\
\hline
\rowcolor[gray]{0.96}
\textbf{VNet} & 62.88 $\pm$ 3.92 & 45.95 $\pm$ 4.28 & 46.17 $\pm$ 4.70 & 69.21 $\pm$ 1.51 & 9.78 $\pm$ 1.76 & 3.64 $\pm$ 0.55 \\

\rowcolor[gray]{0.96}
\textbf{SegResNet} & 64.01 $\pm$ 11.59 & 49.10 $\pm$ 14.78 & 59.27 $\pm$ 7.51 & 70.92 $\pm$ 17.13 & 10.96 $\pm$ 2.97 & 3.03 $\pm$ 1.37 \\

\rowcolor[gray]{0.96}
\textbf{DynUNet} & 62.27 $\pm$ 8.14 & 45.62 $\pm$ 9.13 & 65.68 $\pm$ 5.80 & 59.34 $\pm$ 10.22 & 8.59 $\pm$ 1.57 & 1.92 $\pm$ 0.30 \\
\hline
\rowcolor[gray]{0.96}
\textbf{UNETR} & 72.19 $\pm$ 4.93 & 56.65 $\pm$ 5.82 & 63.74 $\pm$ 5.30 & 86.73 $\pm$ 18.42 & 13.06 $\pm$ 10.38 & 2.87 $\pm$ 0.93 \\

\rowcolor[gray]{0.96}
\textbf{SwinUNETR} & 76.65 $\pm$ 9.04 & 62.75 $\pm$ 11.13 & \textbf{76.09 $\pm$ 0.83} & 74.70 $\pm$ 16.40 & 9.80 $\pm$ 12.13 & 2.05 $\pm$ 1.78 \\
\hline
\rowcolor[gray]{0.96}
\textbf{Ours} & \textbf{77.11 $\pm$ 3.57} & \textbf{63.23 $\pm$ 4.52} & 75.93 $\pm$ 5.18 & \textbf{75.17 $\pm$ 8.87} & \textbf{5.70 $\pm$ 0.63} & \textbf{1.39 $\pm$ 0.22} \\
\hline
\end{tabular}
\end{table*}

\section{Experiments}
\subsection{Dataset and Implementation}
In this study, we retrospectively reviewed patients who underwent proton radiotherapy at Mayo Clinic Arizona between 2016 and 2025, and established a head-and-neck radiotherapy-induced normal tissue injury dataset by screening cases through case-by-case review of medical records and follow-up imaging. The final dataset comprised 70 cases in total, including 29 cases of osteoradionecrosis (ORN), 19 cases of cerebral edema (CE), and 22 cases of cerebral radiation necrosis (CRN). The injury regions are manually delineated by clinicians on the imaging modality most appropriate for each injury type, with ORN annotated on CT, CE on T2-weighted or T2-FLAIR MR images, and CRN on contrast-enhanced T1-weighted MR images. In this way, each injury type is labeled on the modality that best captures its pathological characteristics, providing clinically meaningful voxel-level annotations for model development and evaluation. Furthermore, to enable the use of radiotherapy dose information in our framework, the plan CT is registered to the corresponding image used for lesion delineation, and the associated dose distribution map is subsequently aligned to the same image space. This process allows the spatially matched dose map to be used together with the annotated images in our algorithm.

To ensure balanced evaluation across injury types, the dataset is split at the case level within each category using an approximately 4:1 ratio for training and testing. Given the limited size of the training data, we further apply a series of data augmentation strategies during training to improve model robustness, including random flipping, random affine transformation, random noise, random blur, and random gamma augmentation. Before training, the input image, lesion mask, and high-dose ROI mask are resampled to an isotropic voxel spacing of $1\times1\times1~\mathrm{mm}^3$, and the input image is subsequently normalized using Z-score normalization over nonzero voxels. The model is implemented in PyTorch and trained on two GPUs. Optimization is performed using AdamW with $\beta_1=0.9$, $\beta_2=0.999$, and a weight decay of $0.01$. The initial learning rate is set to $3\times10^{-4}$ and is decayed by a factor of $0.1$ using a MultiStepLR scheduler. The model is trained for 100 epochs with a batch size of 4, while the click-based refinement process is performed for 3 iterations, with 4 simulated clicks incorporated at each iteration.

\subsection{Evaluation Metrics}
We use six evaluation metrics, including Dice, IoU, Precision, Recall, HD95, and ASSD, to comprehensively assess segmentation performance, and these metrics can be broadly grouped into two categories, i.e., overlap-based metrics and boundary-distance-based metrics. Among them, Dice, IoU, Precision, and Recall are adopted to evaluate the degree of overlap between the predicted mask and the ground-truth mask. Among them, Dice and IoU measure the overall spatial agreement between the two masks, while Precision and Recall further characterize false-positive and false-negative tendencies, respectively. For these overlap-based metrics, higher values indicate better segmentation performance.

To further assess boundary accuracy, we additionally use two boundary-distance-based metrics, namely the 95th percentile Hausdorff Distance (HD95) and the Average Symmetric Surface Distance (ASSD), to evaluate the geometric discrepancy between the predicted boundary and the ground-truth boundary. HD95 measures the 95th percentile of the surface distances between the two masks and is less sensitive to extreme outliers than the conventional Hausdorff Distance, whereas ASSD reflects the average bidirectional surface distance and provides a more stable assessment of overall boundary accuracy. For both HD95 and ASSD, lower values indicate better boundary agreement.

\begin{figure*}[!t]
\centering
\begin{overpic}[width=1\linewidth]{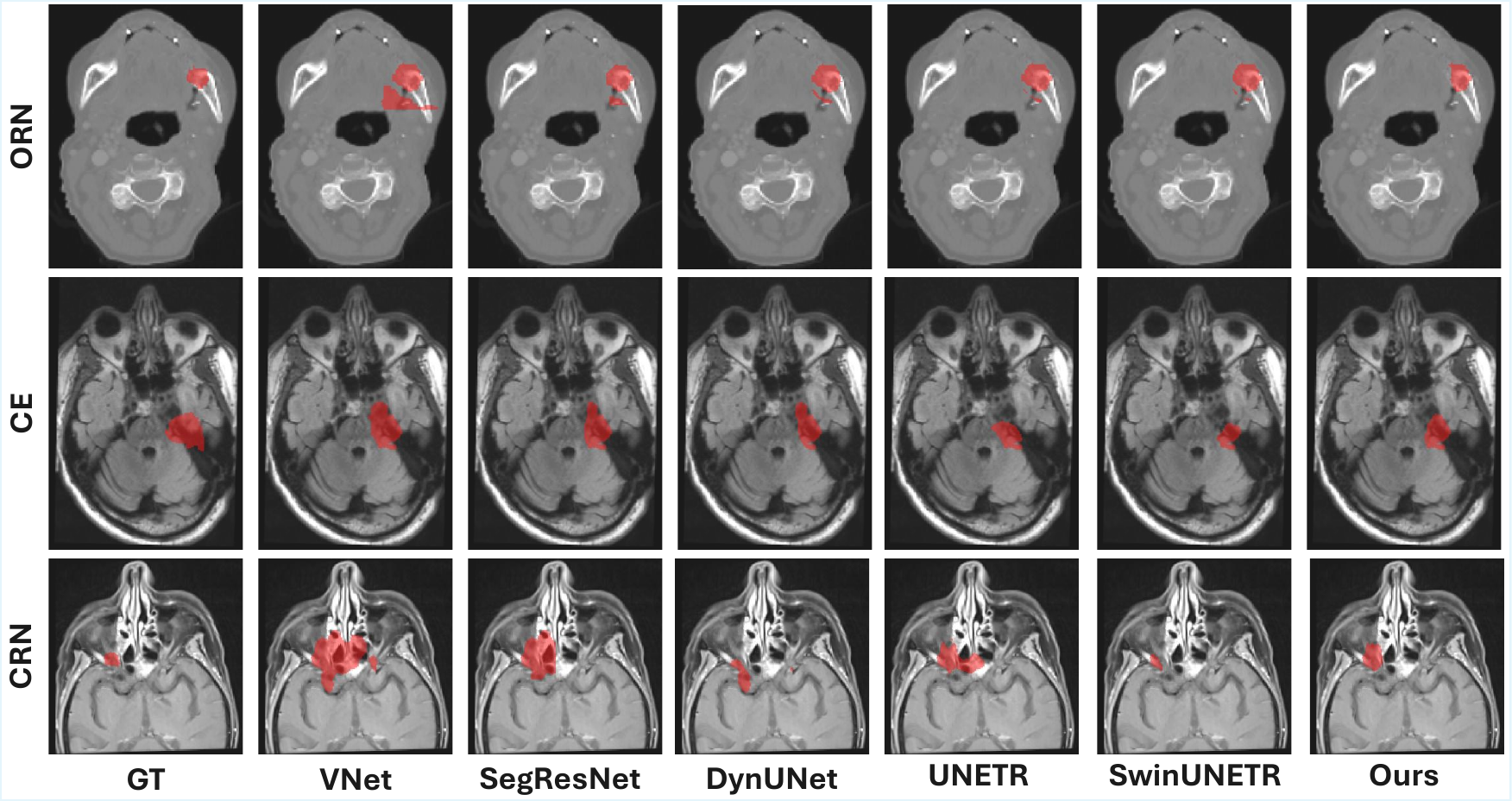}
    \end{overpic}
    \vspace{-4mm}
\centering
\caption{Visual comparison of segmentation results produced by different methods for three representative cases of radiotherapy-induced injuries, including osteoradionecrosis (ORN), cerebral edema
(CE), cerebral radiation necrosis (CRN). From left to right are the ground truth (GT), and the segmentation results obtained by VNet, SegResNet, DynUNet, UNETR, SwinUNETR and our method. Red overlays indicate the lesion regions.}
\label{comparision}
\end{figure*}

\subsection{Comparison with State-of-the-art Methods}
We compare the proposed method with five state-of-the-art medical image segmentation methods, including three CNN-based methods, i.e., VNet~\citep{milletari2016vnet}, SegResNet~\citep{myronenko20183d} and DynUNet~\citep{isensee2021nnu}, as well as two Transformer-based methods, i.e., UNETR~\citep{hatamizadeh2022unetr} and SwinUNETR~\citep{hatamizadeh2022swin}. Specifically, VNet is a classical volumetric segmentation network based on an encoder-decoder architecture with 3D convolutions, SegResNet introduces residual learning into 3D medical image segmentation, and DynUNet adopts a dynamically configured U-Net architecture for enhanced multiscale representation learning. In contrast, UNETR employs a Transformer encoder to capture long-range dependencies in volumetric images, while SwinUNETR further incorporates hierarchical Swin Transformer blocks for more effective local-global feature modeling. To ensure a fair comparison, all methods are trained and evaluated under the same data split and experimental settings. The quantitative and qualitative results are presented in Table~\ref{tab:sota_comparison} and Fig.~\ref{comparision}, respectively.

\subsubsection{Quantitative Comparison}
The quantitative results in Table~\ref{tab:sota_comparison} show that the proposed method achieves the best overall performance across most evaluation metrics. It yields the highest Dice ($77.11\%$), IoU ($63.23\%$) and Recall ($75.17\%$), together with the lowest HD95 ($5.70$) and ASSD ($1.39$). Relative to the strongest competing method, SwinUNETR, the proposed method further improves Dice from $76.65\%$ to $77.11\%$, IoU from $62.75\%$ to $63.23\%$, and Recall from $74.70\%$ to $75.17\%$, while markedly reducing HD95 from $9.80$ to $5.70$ and ASSD from $2.05$ to $1.39$. The especially large gains in boundary-distance-based metrics indicate that the proposed framework improves not only lesion overlap but also boundary delineation, yielding more spatially accurate and consistent predictions. This pattern is in line with the coarse-to-fine design of our method, where large-scale pretraining and progressive prompting jointly facilitate lesion localization and refinement. It is also notable that SwinUNETR attains the highest Precision ($76.09\%$), whereas our method remains comparable in Precision ($75.93\%$) while outperforming it on all other metrics. This suggests that the proposed method achieves a better overall balance between false-positive control and lesion coverage, which is particularly relevant for radiotherapy-induced injuries with small, sparse and irregular appearances. Moreover, the inferior performance of VNet, SegResNet and DynUNet relative to UNETR and SwinUNETR suggests the importance of global contextual modeling for this task. Our method further builds on this advantage by incorporating task-aware prompting and adaptation on top of large-scale pretrained representations.

\subsubsection{Qualitative Comparison}
The qualitative results in Fig.~\ref{comparision} further support the quantitative findings. For the ORN case, VNet produces clear over-segmentation, whereas SegResNet and DynUNet still show inaccurate estimation of lesion extent. UNETR and SwinUNETR generate more reasonable predictions, but their results remain less consistent with the ground truth than those of the proposed method. By contrast, our method produces a lesion shape and extent that are visually closest to the ground truth. Similar trends are observed for the CE and CRN cases. In particular, for CRN case, several competing methods produce substantial false-positive regions or distorted lesion morphology, whereas the proposed method better preserves the true lesion location and overall shape. These observations are consistent with the design of our framework. The progressive prompting strategy provides complementary guidance for task adaptation, lesion localization and iterative refinement, which is particularly beneficial for radiotherapy-induced injuries with small, sparse and irregular appearances. In combination with large-scale medical image pretraining, this design helps the model focus on clinically relevant regions and produce predictions that are more accurate in both lesion extent and boundary morphology.

\begin{table*}[!t]
\setlength{\abovecaptionskip}{0.1cm}
\setlength{\belowcaptionskip}{0.1cm}
\centering
\renewcommand\arraystretch{1.25}
\setlength{\tabcolsep}{6pt}
\arrayrulecolor{black}
\caption{Quantitative ablation study of the progressively introduced components in the proposed framework for radiotherapy-induced injury segmentation, evaluated using Dice, IoU, Precision, Recall, HD95 and ASSD.}
\label{tab:ablation_prompting}
\begin{tabular}{l|c|c|c|c|c|c}
\hline
\rowcolor[gray]{0.85}
\textbf{Method} & \textbf{Dice\%$\uparrow$} & \textbf{IoU\%$\uparrow$} & \textbf{Precision\%$\uparrow$} & \textbf{Recall\%$\uparrow$} & \textbf{HD95$\downarrow$} & \textbf{ASSD$\downarrow$} \\
\hline
\rowcolor[gray]{0.96}
\textbf{Base} & 65.99 $\pm$ 6.46 & 49.08 $\pm$ 6.07 & 51.32 $\pm$ 2.94 & 62.99 $\pm$ 7.14 & 10.59 $\pm$ 1.86 & 1.70 $\pm$ 0.06 \\
\hline
\rowcolor[gray]{0.96}
\textbf{Base-T} & 71.00 $\pm$ 3.31 & 55.11 $\pm$ 3.88 & 62.00 $\pm$ 6.48 & \underline{73.66 $\pm$ 16.99} & 7.77 $\pm$ 1.26 & 2.36 $\pm$ 0.31 \\

\rowcolor[gray]{0.96}
\textbf{Base-T-B} & 72.48 $\pm$ 8.36 & \underline{58.48 $\pm$ 12.53} & 69.44 $\pm$ 6.82 & 68.67 $\pm$ 24.59 & 7.71 $\pm$ 1.92 & 1.83 $\pm$ 0.40 \\

\rowcolor[gray]{0.96}
\textbf{Base-T-B-C} & \underline{74.05 $\pm$ 6.95} & 57.91 $\pm$ 9.12 & \textbf{78.79 $\pm$ 3.39} & 72.76 $\pm$ 15.54 & \underline{6.43 $\pm$ 1.35} & \textbf{1.28 $\pm$ 0.04} \\
\hline
\rowcolor[gray]{0.96}
\textbf{Base-T-B-C-L} & \textbf{77.11 $\pm$ 3.57} & \textbf{63.23 $\pm$ 4.52} & \underline{75.93 $\pm$ 5.18} & \textbf{75.17 $\pm$ 8.87} & \textbf{5.70 $\pm$ 0.63} & \underline{1.39 $\pm$ 0.22} \\
\hline
\end{tabular}
\end{table*}

\begin{figure*}[!t]
\centering
\begin{overpic}[width=1\linewidth]{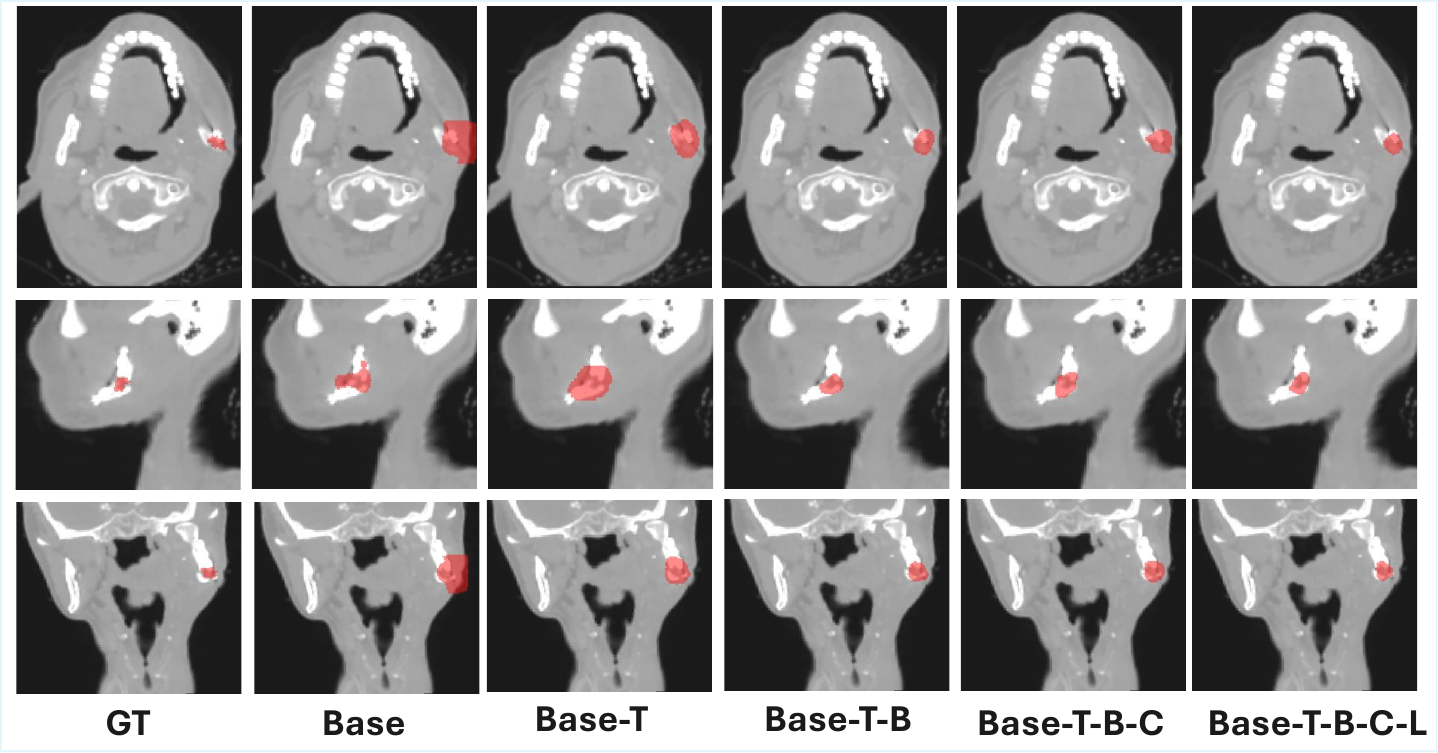}

    \end{overpic}
    \vspace{-6mm}
\centering

\caption{Visual comparison of ablation results for an ORN case in three anatomical views (axial, sagittal and coronal). From left to right are the ground truth (GT) and the results of the baseline model and its successive variants with text prompt (Base-T), text + box prompts (Base-T-B), text + box + click prompts (Base-T-B-C), and the full model with the combined loss (Base-T-B-C-L). Red overlays denote the lesion regions.}
\label{ablation}
\end{figure*}

\subsection{Ablation Study of Key Components}
To verify the contribution of each key component in the proposed framework, we perform an ablation study by progressively introducing the text prompt, box prompt, click prompt and \textit{Small-target Focus Loss}. Specifically, we compare the following five variants: \uppercase\expandafter{\romannumeral1}) \textbf{Base}, the baseline segmentation model without the progressive prompting strategy or small-target focus loss; \uppercase\expandafter{\romannumeral2}) \textbf{Base-T}, which introduces the text prompt for task-aware adaptation; \uppercase\expandafter{\romannumeral3}) \textbf{Base-T-B}, which further incorporates the dose-guided box prompt for coarse lesion localization; \uppercase\expandafter{\romannumeral4}) \textbf{Base-T-B-C}, which further introduces click prompts for iterative lesion refinement; and \uppercase\expandafter{\romannumeral5}) \textbf{Base-T-B-C-L}, the full model, which additionally employs small-target focus loss. All variants are trained and evaluated under the same experimental settings. The quantitative and qualitative results are presented in Table~\ref{tab:ablation_prompting} and Fig.~\ref{ablation}, respectively.

Starting from the baseline model, introducing the text prompt leads to a clear performance improvement. As shown in Table~\ref{tab:ablation_prompting}, Dice increases from $65.99\%$ to $71.00\%$, IoU from $49.08\%$ to $55.11\%$, and HD95 decreases from $10.59$ to $7.77$. These improvements suggest that the text prompt provides useful task-aware semantic guidance for adapting the model to different radiotherapy-induced injury types. This effect is also reflected in Fig.~\ref{ablation}, where the baseline model shows obvious over-segmentation, whereas Base-T produces predictions that are more concentrated around the lesion region, although inaccurate lesion extent and false-positive regions still remain.

Based on Base-T, further introducing the dose-guided box prompt brings additional improvement. Quantitatively, Dice increases to $72.48\%$ and IoU to $58.48\%$, indicating that the box prompt helps the model focus more effectively on lesion-related regions. This trend is visually consistent with Fig.~\ref{ablation}, where Base-T-B shows more accurate coarse localization than Base-T in all three views, with fewer irrelevant responses outside the lesion region. However, the predicted region remains relatively coarse, suggesting that box prompting mainly benefits lesion localization rather than fine-grained boundary delineation.

When click prompting is further incorporated, the most notable improvement appears in Precision, HD95, and ASSD, with Precision rising markedly from $69.44\%$ to $78.79\%$, HD95 decreasing from $7.71$ to $6.43$, and ASSD decreasing from $1.83$ to $1.28$. Dice also increases from $72.48\%$ to $74.05\%$, while IoU remains largely comparable ($58.48\%$ vs. $57.91\%$). These changes indicate that click prompting is particularly effective for suppressing false-positive predictions and refining local boundary details, while also bringing a moderate gain in lesion coverage. This role is also evident in Fig.~\ref{ablation}, where Base-T-B-C produces a lesion shape that is more compact and visually closer to the ground truth than Base-T-B, especially in the axial and coronal views. Such observations are consistent with the intended role of click prompting as an iterative refinement mechanism for small and irregular lesions.

\begin{figure}[!t]
\setlength{\belowcaptionskip}{-0.4cm}
\centering
\begin{overpic}[width=1\linewidth]{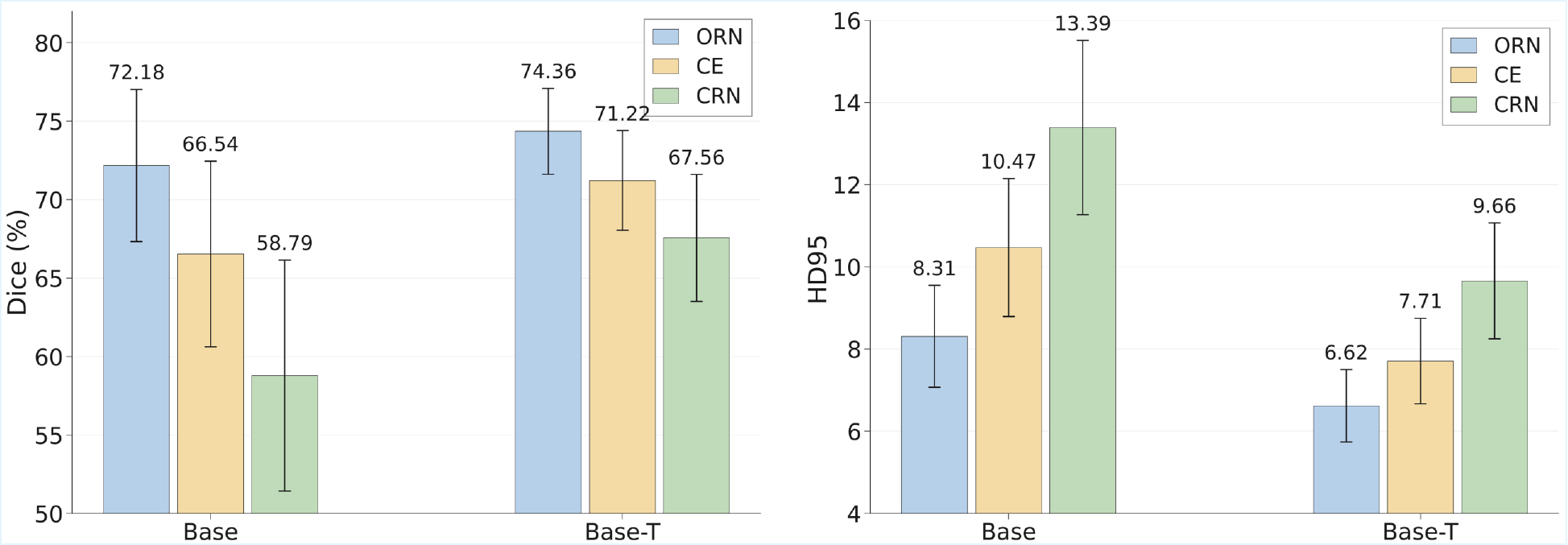}

    \end{overpic}
    \vspace{-1mm}
\centering
 \setlength{\abovecaptionskip}{0.1cm}
 \vspace{-2mm}
\caption{Per-task comparison of Base and Base-T on ORN, CE, and CRN for evaluating the contribution of text prompting to unified multi-task segmentation. The left panel shows Dice, and the right panel shows HD95.}
\label{text_mutil_task}
\end{figure}

\begin{figure*}[!t]
\setlength{\belowcaptionskip}{-0.4cm}
\centering
\begin{overpic}[width=1\linewidth]{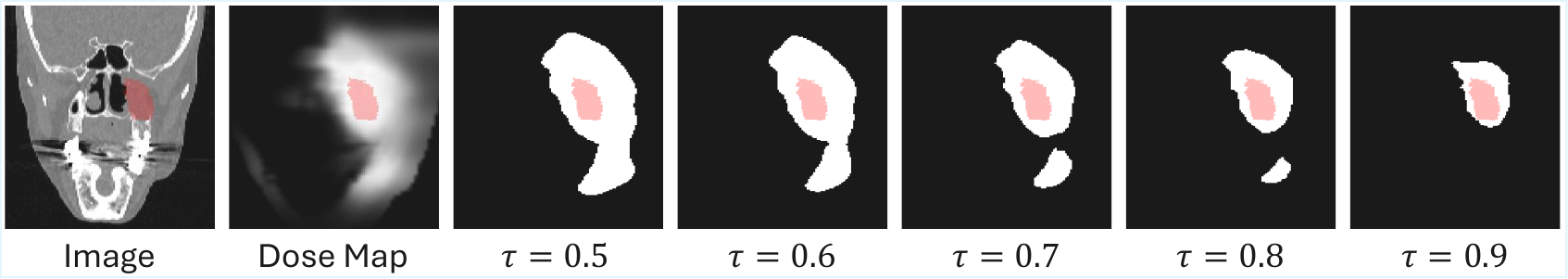}

    \end{overpic}
    \vspace{-1mm}
\centering
 \setlength{\abovecaptionskip}{0.1cm}
 \vspace{-2mm}
\caption{Visualization of high-dose masks extracted under different relative dose thresholds for dose-guided box prompting in an ORN case. From left to right are the input image, the corresponding dose map, and the high-dose masks obtained using different relative thresholds $(\tau \times D_{\max})$, with $\tau = 0.5, 0.6, 0.7, 0.8,$ and $0.9$. The lesion region is overlaid in red.}
\label{high-dose mask}
\end{figure*}

Finally, adding the small-target focus loss further improves the overall segmentation performance and yields the full model. Compared with Base-T-B-C, the full model increases Dice from $74.05\%$ to $77.11\%$, IoU from $57.91\%$ to $63.23\%$, and Recall from $72.76\%$ to $75.17\%$, while further reducing HD95 from $6.43$ to $5.70$. In Fig.~\ref{ablation}, the full model produces the segmentation result that is visually closest to the ground truth across axial, sagittal and coronal views, with more accurate lesion extent and cleaner boundaries. These results suggest that the small-target focus loss further enhances the model's sensitivity to small lesion regions and improves the balance between lesion coverage and boundary precision.

Overall, the ablation study shows that the proposed components are complementary. The text prompt primarily facilitates task-aware adaptation, the box prompt improves coarse lesion localization, the click prompt refines local predictions and suppresses false positives, and the small-target focus loss further strengthens segmentation of small-target regions. Their combination leads to the best overall performance.

\section{DISCUSSION}
\subsection{Contribution of Text Prompting to Unified Multi-Task Segmentation}
To further investigate the role of text prompting in unified multi-task segmentation, we conduct an additional experiment to evaluate whether the introduced text prompt can provide effective task-aware guidance across different injury types. Specifically, we compare the baseline model without text prompting (Base) and the model with text prompting (Base-T), while keeping all other settings unchanged. The comparison is performed separately on the three injury segmentation tasks, including osteoradionecrosis (ORN), cerebral edema (CE), and cerebral radiation necrosis (CRN). Considering that Dice reflects overall overlap accuracy and HD95 measures boundary discrepancy, we report these two metrics for per-task analysis, as shown in Fig.~\ref{text_mutil_task}.

As shown in Fig.~\ref{text_mutil_task}, introducing text prompting consistently improves segmentation performance across all three injury types. For ORN, Dice increases from 72.18\% to 74.36\%, while HD95 decreases from 8.31 to 6.62. For CE, Dice improves from 66.54\% to 71.22\%, with HD95 reduced from 10.47 to 7.71. The largest improvement is observed for CRN, where Dice rises from 58.79\% to 67.56\% and HD95 decreases from 13.39 to 9.66. These results indicate that text prompting provides effective task-aware guidance within the unified framework. Notably, the improvement is more pronounced for CE and especially CRN, which are relatively more challenging than ORN. This suggests that text prompting is particularly beneficial when the task becomes more complex and heterogeneous. By explicitly encoding task-related semantic information, the text prompt helps the model better distinguish different injury manifestations and adapt its segmentation behavior within a shared model.

\subsection{Effect of Threshold Selection on Dose-guided Box Prompting}
As described in Section 3.4, in the dose-guided box prompting strategy, we extract a high-dose mask from the dose distribution map using a threshold $\tau$, and then generate the dose-guided box from this mask to provide spatial guidance for segmentation. In this section, we further investigate the effect of threshold selection on dose-guided box prompting. We first visualize the high-dose masks obtained under different threshold settings, as shown in Fig.~\ref{high-dose mask}. As the threshold increases from $\tau=0.5$ to $\tau=0.9$, the extracted high-dose region gradually shrinks and becomes more concentrated around the dose peak. Accordingly, lower thresholds produce a broader region and thus a looser spatial prior, whereas higher thresholds yield a more compact region, providing a tighter, more focused localization cue.

\begin{figure}[!t]
\setlength{\belowcaptionskip}{-0.4cm}
\centering
\begin{overpic}[width=1\linewidth]{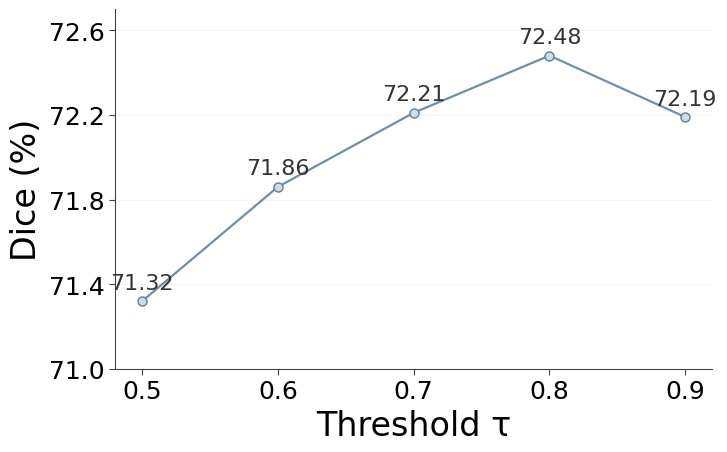}

    \end{overpic}
    \vspace{-1mm}
\centering
 \setlength{\abovecaptionskip}{0.1cm}
 \vspace{-2mm}
\caption{Quantitative analysis of segmentation performance under different relative dose thresholds for high-dose mask extraction in dose-guided box prompting.}
\label{dose-guide-prompt}
\end{figure}

We further evaluate the final segmentation performance under different threshold settings using Dice, as shown in Fig.~\ref{dose-guide-prompt}. The Dice score gradually increases from $71.32\%$ at $\tau=0.5$ to $72.48\%$ at $\tau=0.8$, followed by a slight decrease to $72.19\%$ at $\tau=0.9$. Overall, the variation remains limited, particularly from $\tau=0.7$ to $\tau=0.9$, where the Dice scores are highly comparable. These results suggest that the proposed dose-guided box prompting strategy is reasonably robust to threshold selection within a practical range. This trend is also consistent with the role of the dose-guided box prompt in our framework. A relatively low threshold yields an overly large high-dose region, resulting in a less focused coarse localization prior. As the threshold increases, the spatial prior becomes more compact and informative, leading to improved segmentation performance. When the threshold becomes too high, however, the extracted region may become slightly over-concentrated and exclude part of the clinically relevant surrounding area, resulting in a small performance drop. Overall, the limited variation across thresholds indicates that the proposed framework does not rely heavily on a single hand-crafted threshold.

\subsection{Sensitivity Analysis of Click-based Refinement}
We further investigate the proposed click-based refinement strategy through a sensitivity analysis of two key factors, i.e., the number of refinement iterations and the number of simulated clicks per iteration. To isolate the effect of click-based refinement itself, this analysis is conducted on the Base-T-B variant, which already incorporates the text prompt for task conditioning and the dose-guided box prompt for coarse lesion localization, but does not yet include click prompting or the small-target focus loss. We choose this setting because it provides a suitable basis for evaluating click-based refinement after semantic guidance and coarse spatial localization have already been established, while avoiding additional interference from the loss design. Since the two click-related variables jointly affect the refinement process, determining a strictly optimal combination would require a much larger number of experiments through exhaustive search over different parameter combinations. Instead, to obtain a relatively suitable setting under controllable experimental cost, we adopt a stepwise sensitivity analysis strategy, in which one factor is varied while the other is kept fixed.

\begin{figure}[!t]
\setlength{\belowcaptionskip}{-0.4cm}
\centering
\begin{overpic}[width=1\linewidth]{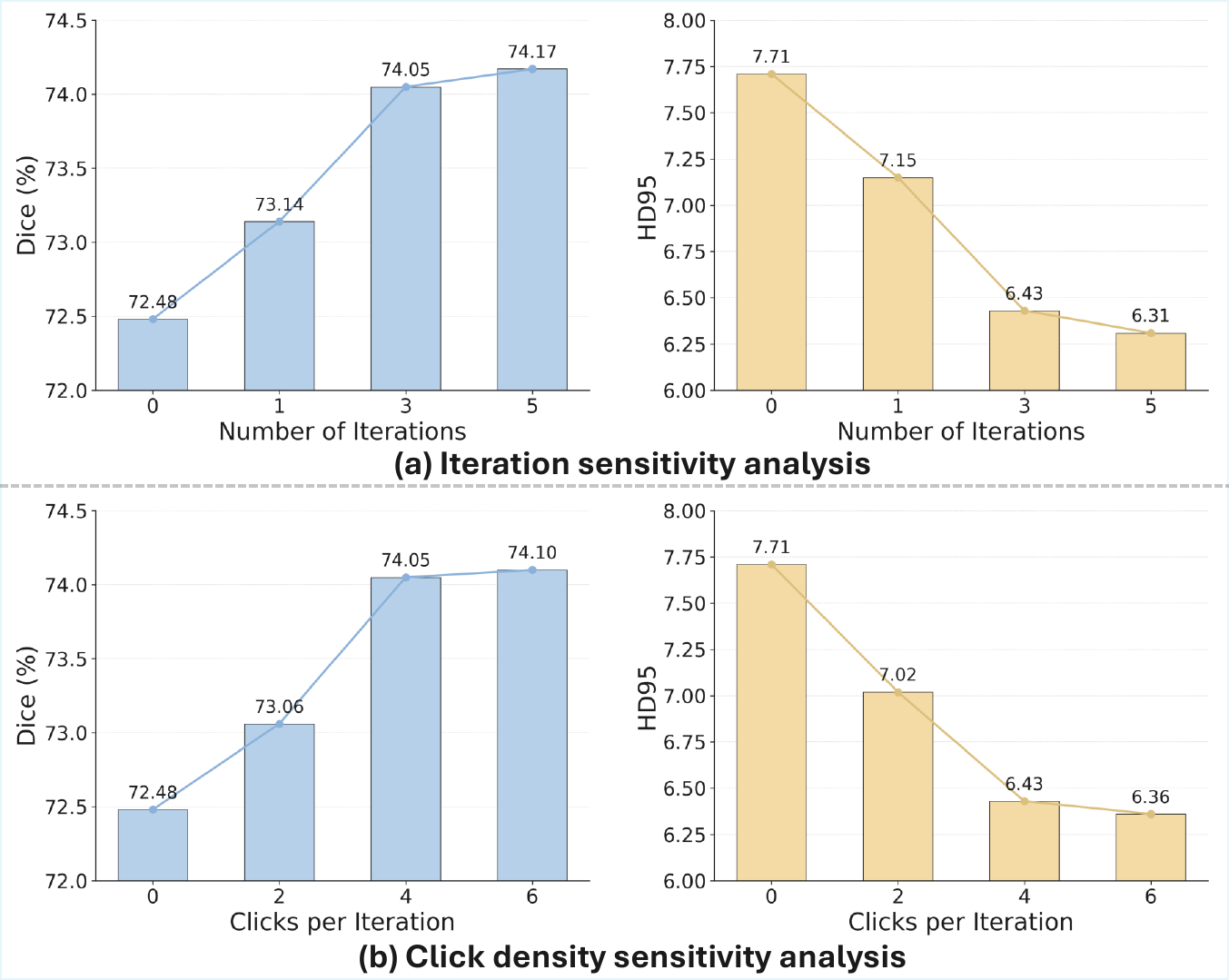}

    \end{overpic}
    \vspace{-1mm}
\centering
 \setlength{\abovecaptionskip}{0.1cm}
 \vspace{-2mm}
\caption{Sensitivity analysis of the proposed click refinement strategy in terms of Dice and HD95 under different refinement iteration settings and simulated click numbers. (a) Quantitative results with different numbers of refinement iterations. (b) Quantitative results with different numbers of simulated clicks per iteration.}
\label{click_anaylsis}
\end{figure}

Specifically, we first analyze the effect of refinement iteration number. To isolate this factor, the number of simulated clicks per iteration is fixed, with foreground and background clicks equally divided. Based on our prior observation that radiotherapy-induced injury regions are usually very small, we choose a relatively small click number for sensitivity analysis. Accordingly, the number of simulated clicks per iteration is set to 4. Under this setting, we compare different numbers of refinement iterations and evaluate the results using Dice and HD95, as shown in Fig.~\ref{click_anaylsis}(a). The results show that increasing the number of iterations consistently improves segmentation performance. Dice increases from 72.48\% at 0 iterations to 73.14\% at 1 iteration and further to 74.05\% at 3 iterations, while HD95 decreases from 7.71 to 7.15 and then to 6.43. When the number of iterations is further increased to 5, the improvement becomes marginal, with Dice reaching 74.17\% and HD95 decreasing slightly to 6.31. These findings indicate that iterative refinement is effective for progressively correcting local prediction errors, while most of the performance gain can already be achieved within 3 iterations. Considering both performance and efficiency, we therefore select 3 as the refinement iteration number for the subsequent analysis.

Based on the selected iteration number, we then analyze the effect of the number of simulated clicks per iteration by fixing the number of refinement iterations at 3 and varying the click number, as shown in Fig.~\ref{click_anaylsis}(b). The results show that increasing the number of simulated clicks leads to clear performance improvement at the early stage. Compared with the setting without click refinement, using 2 clicks per iteration improves Dice from 72.48\% to 73.06\% and reduces HD95 from 7.71 to 7.02. Increasing the click number to 4 further improves Dice to 74.05\% and reduces HD95 to 6.43. However, when the click number is further increased to 6, the additional gain becomes very limited, with Dice only slightly increasing to 74.10\% and HD95 slightly decreasing to 6.36. These results support the reasonableness of our empirical choice of using a relatively small click number and further indicate that, under 3 refinement iterations, 4 clicks per iteration provides a relatively favorable setting among the tested options. 

Overall, the sensitivity analysis shows that both refinement iteration number and simulated click number influence the effectiveness of click refinement, while their benefits gradually saturate beyond a moderate range. Based on these results, we use 3 refinement iterations and 4 simulated clicks per iteration in the main experiments, as these values provide a relatively favorable balance between segmentation performance and experimental efficiency among the tested settings, rather than a strictly optimal combination obtained from exhaustive search.

\section{Conclusion}
In this paper, we investigate automatic segmentation of radiotherapy-induced normal tissue injuries and develop a novel 3D SAM-based progressive prompting framework for unified multi-task segmentation under limited-data conditions, leveraging large-scale pretrained priors from the SAM framework to alleviate the dependence on task-specific annotated data. To address the challenges of scarce annotations, task heterogeneity, and the extremely small and sparse nature of injury regions, we introduce several targeted designs into the framework, including 1) text prompts to provide task-aware semantic guidance for unified segmentation of different injury types, 2) dose-guided box prompts to exploit radiotherapy priors for coarse lesion localization, 3) click-based refinement to progressively correct local prediction errors, and 4) a small-target focus loss to strengthen supervision on clinically relevant regions. Extensive experiments demonstrate that our method consistently outperforms state-of-the-art segmentation approaches and achieves reliable performance across ORN, CE, and CRN, highlighting its potential for future radiotherapy toxicity assessment and longitudinal monitoring.




\section*{Declaration of Competing Interest}
The authors declare that they have no known competing financial interests or personal relationships that could have appeared to influence the work reported in this paper.



\section*{Acknowledgments}
This work was supported in part by the Department of Radiation Oncology, Mayo Clinic, Rochester, Minnesota, and Phoenix, Arizona, USA, R01 CA 261932 and R01 CA 272602 to Robert W. Mutter, and P30 CA015083 to MCCCC. This research was also supported by NIH/BIBIB R01EB293388, by NIH/NCI R01CA280134, by the Eric \& Wendy Schmidt Fund for AI Research
\& Innovation, and by the Kemper Marley Foundation to Wei Liu.

\bibliographystyle{model2-names}
\biboptions{authoryear}
\bibliography{refs}


\end{document}